\documentclass[10pt,twocolumn,letterpaper]{article}

\usepackage{iccv}
\usepackage{times}
\usepackage{epsfig}
\usepackage{graphicx}
\usepackage{amsmath}
\usepackage{amssymb}

\usepackage{booktabs}
\usepackage{array, multirow, boldline}
\usepackage{tabularx}
\usepackage{subcaption}
\usepackage{adjustbox}
\usepackage{wrapfig}
\usepackage{caption}
\usepackage{tikz}
\usepackage[breaklinks=true,bookmarks=false]{hyperref}
\usepackage{cleveref}
\iccvfinalcopy 

\def\iccvPaperID{****} 

\ificcvfinal\fi

\begin{document}

\title{A SAM-based Solution for Hierarchical Panoptic Segmentation of Crops and Weeds Competition}

\author{Khoa Dang Nguyen$^*$\\ \tt\small{{khoanguyen.de11@nycu.edu.tw}}
\and
Thanh-Hai Phung$^*$\\ \tt\small{{haipt.ee08@nycu.edu.tw}}
\and
Hoang-Giang Cao$^\dagger$\\ \tt\small{{chgiang@cit.ctu.edu.vn}}
\and
National Yang Ming Chiao Tung University\\ Taiwan\\
}



\maketitle

\def\thefootnote{$*$}\footnotetext{Equal contribution.}\def\thefootnote{\arabic{footnote}}
\def\thefootnote{$\dagger$}\footnotetext{Correspondence.}\def\thefootnote{\arabic{footnote}}

\begin{abstract}
Panoptic segmentation in agriculture is an advanced computer vision technique that provides a comprehensive understanding of field composition.
It facilitates various tasks such as crop and weed segmentation, plant panoptic segmentation, and leaf instance segmentation, all aimed at addressing challenges in agriculture.
Exploring the application of panoptic segmentation in agriculture, the 8th Workshop on Computer Vision in Plant Phenotyping and Agriculture (CVPPA) hosted the challenge of hierarchical panoptic segmentation of crops and weeds using the PhenoBench dataset.
To tackle the tasks presented in this competition, we propose an approach that combines the effectiveness of the Segment AnyThing Model (SAM) for instance segmentation with prompt input from object detection models.
Specifically, we integrated two notable approaches in object detection, namely DINO and YOLO-v8.
Our best-performing model achieved a PQ+ score of 81.33 based on the evaluation metrics of the competition.
\end{abstract}

\section{Introduction}
Deep learning has revolutionized the field of computer vision by enabling machines to interpret and understand visual data with remarkable achievements. In agriculture, computer vision powered by deep learning has a cutting-edge technique called Panoptic Segmentation.
Panoptic segmentation is a recent advent task of computer vision that combines both semantic segmentation and instance segmentation to provide a comprehensive understanding of the agricultural landscape.

To further explore the advantages of panoptic segmentation in agriculture, the $8^{th}$ Workshop on Computer Vision in Plant Phenotyping and Agriculture (CVPPA) hosted the challenge of Hierarchical Panoptic Segmentation of Crops and Weeds using PhenoBench dataset \cite{weyler2023phenobench}.
The Phenobench dataset is an extensive collection of high-quality images from agricultural fields with detailed annotations.
This dataset supports the tasks that cover semantic segmentation of crops and weeds, panoptic segmentation of plants, leaf instance segmentation, and detection of plants and leaves.
The challenge requires the model to accurately identify and differentiate between sugar beets and weeds in agricultural images.
It involves hierarchical segmentation to distinguish object parts and assign unique labels to each sugar beet or weed instance within the image.

In this paper, we will report on our solution for the competition.
Our approach combines recent models in instance segmentation and object detection.
Specifically, we employ the Segmentation Anything Model (SAM) for instance segmentation, and for object detection, we utilize YOLO-v8 and DINO.
We apply different fine-tuning techniques to individually improve the performance of each model on their respective tasks.
Subsequently, we integrate these fine-tuned models into a pipeline, shown in \Cref{fig:pipeline}.
As illustrated in the pipeline, SAM-based segmentation takes the bounding box generated by the object detection model as its prompt to produce instance segmentation predictions for a given image.

\section{Our Approach}

In this section, we introduce our approach in detail.
With the goal of providing a comprehensive understanding of the scene by segmenting the image into semantically meaningful parts or regions while also detecting and distinguishing individual instances of objects within those regions.

\begin{figure}[t]
\begin{center}
        \includegraphics[width=0.7\linewidth]{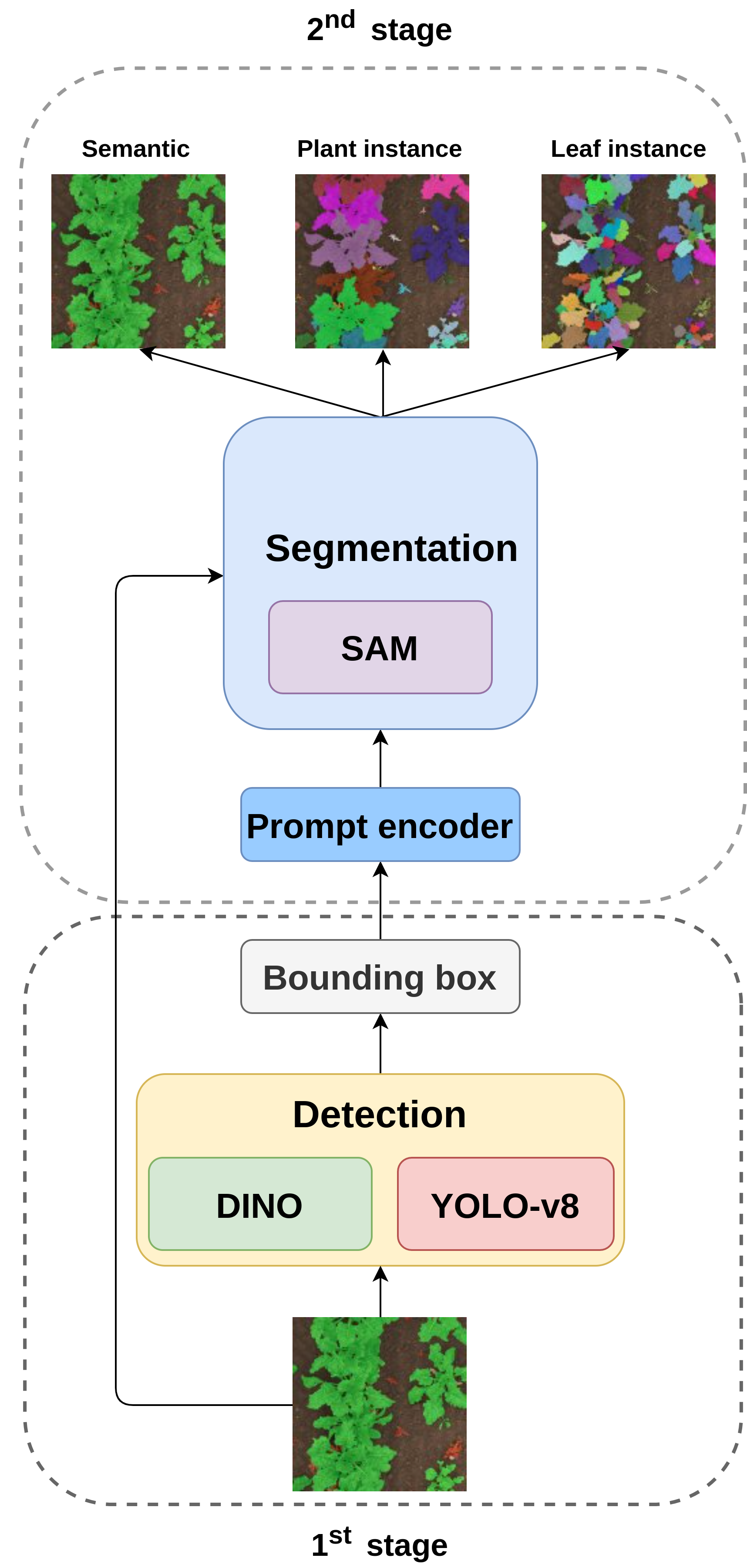}
        \caption{Pipeline of our approach combining SAM-based for segmentation and prompt input generated from object detection models (YOLO-v8 and DINO).
        }
  \label{fig:pipeline}
\end{center}  
\end{figure}

Our approach capitalizes on the effectiveness of SAM-based models for segmentation, with input prompts generated by object detection models, typically in the form of bounding boxes.
So, our pipeline is structured into two modules: SAM-based segmentation and object detection (illustrated in \Cref{fig:pipeline}).
For segmentation using SAM-based models, we employ the Segment Anything in High-Quality (HQ-SAM) model \cite{ke2023segment}.
For object detection, our choices include DINO and YOLO-v8.
Subsequently, we fine-tune each module with the Phenobench dataset.
It's worth noting that the training of these two modules can be processed concurrently in parallel.


In general, our training configuration is as follows:
Our proposed network is implemented in PyTorch. 
We use the AdamW optimizer with a momentum of 0.9 and a weight decay of 1e-4 for optimization.
The learning rate is adjusted by a factor of 1e-5 following a learning rate decay of 0.9.
During training, we employ a Cosine annealing schedule, which adapts to the variations in data and monitors the validation mean Intersection over Union (IoU) after each epoch.
We apply an early stopping technique if the mean IoU metric does not improve after 20 epochs.
The batch size is set to 4.
Our training server is equipped with 4 NVIDIA GeForce RTX 3090 GPUs.

\subsection{Segmentation}
While the Segment Anything model (SAM) \cite{kirillov2023segment}, trained on a dataset comprising 1.1 billion masks, generally performs well, it often encounters challenges when dealing with objects exhibiting intricate structures or when distinguishing between foreground objects and complex background areas. 
This difficulty is particularly noticeable when handling objects with thin lines,  such as those found in the PhenoBench dataset.

A work \cite{ke2023segment} introduced Segment Anything in High-Quality model (HQ-SAM) with a learnable HQ-Output Token for the mask decoder from intermediate features of the image encoder.
This new design of HQ-SAM improves the quality of mask predictions, particularly for objects with intricate structures, addressing the limitations that SAM faces.
Due to these advantages, we utilize HQ-SAM as our SAM-based segmentation model empirically based on the distribution of the PhenoBench dataset \cite{weyler2023phenobench}.

We use two pre-trained models from HQ-SAM: ViT-Base HQ-SAM and ViT-Huge HQ-SAM.
We fine-tune the HQ-SAM model with PhenoBench training data.
The input to HQ-SAM comprises 'ground-truth' bounding boxes extracted from training data masks.

To enhance the ability of HQ-SAM to adapt to bounding boxes predicted by object detection models, which might not be perfectly precise, we applied box augmentation during fine-tuning.
Box augmentation involves adding random noise ranging from 10\% of the box side length to a maximum of 20 pixels to each coordinate of the ground-truth boxes.

To speed up the training, we also applied LoRA\cite{hu2021lora} during the fine-tuning of HQ-SAM.
LoRA, originally designed for Large Language Models (LLMs), enhances GPU memory efficiency by incorporating memory-saving matrices into pre-trained models.
This allows for fine-tuning on consumer GPUs, therefore improving accessibility and performance.
Building on this concept, a work SonarSAM \cite{wang2023sonarsam} uses LoRA incorporating with SAM-based model for sonar images.
In our solution, we adopted the  LoRA implementation from this work.

LoRA, in our approach, helps to deal with a large-size model like ViT-Huge HQ-SAM, where directly fully fine-tuning this model is impractical.
We apply LoRA for tuning both the image encoder and mask decoder with a rank set as 4, which shows the best performance in our experiments.


\begin{table*}[t]
\caption{Experiment results with different configurations. The best result is highlighted in \textbf{bold}.}
\begin{adjustbox}{width=\linewidth}
\def\arraystretch{1.3}
\begin{tabular}{cccccccccc}
\hline

\multirow{2}{*}{\textbf{Submit time}} & \multirow{2}{*}{\textbf{Date}} & \multicolumn{2}{c}{\textbf{Configurations}} & \multicolumn{6}{c}{\textbf{PhenoBench}} \\ \cline{3-10}
 &                   &  \textbf{Segmentation}                  &  \textbf{Detection}                & \multicolumn{1}{c}{\textbf{IoU (soil)}} & \multicolumn{1}{c}{\textbf{IoU (weed)}} & \multicolumn{1}{c}{\textbf{PQ (leaf)}} & \multicolumn{1}{c}{\textbf{PQ (crop)}} & \multicolumn{1}{c}{\textbf{PQ}} & \textbf{PQ+} \\ \hline
(1)            &  07/14            & SAM (Vit-B)                 &  YOLO-v8 nano           & 99.08      & 61.92     & 63.57     & 69.08     & 66.33 & 73.41  \\
(2)            &  07/27            &  SAM (Vit-B)                 &   YOLO-v8 nano          & 98.95      & 62.71      & 66.39     & 69.52     & 67.95 & 74.39 \\
(3)            &  08/04            &  SAM (Vit-B)  &    YOLO-v8 nano         & 99.21      & 64.70      & 66.49    & 78.35      & 72.42 & 77.19 \\
(4)            &  08/13            &  SAM  (Vit-H) with LoRA  &    YOLO-v8 nano         & 99.13      & 67.78      & 73.95     & 81.02     & 77.48 & 80.47 \\
(5)            &  08/21            &  SAM  (Vit-H) with LoRA  &  DINO        & 99.15      & 70.08      & 73.95     & 81.19     & 77.57 & 81.09 \\ \hline
\textbf{(6)}   & \textbf{08/29}    &  SAM  (Vit-H) with LoRA &    DINO         & \textbf{99.18}     & \textbf{70.66}      & \textbf{73.81}     & \textbf{81.66}     & \textbf{77.73} & \textbf{81.33} \\ \hline
\end{tabular}
\end{adjustbox}
\label{tab:experiment_result}
\end{table*}

\subsection{Object Detection}
Since SAM-like model is a prompt-able segmentation model, the prompt is a must.
We trained an object detector to extract bounding boxes as a prompt for SAM.
For the choice of object detector, we consider between DINO \cite{zhang2022dino} and YOLO-v8 \cite{jocher2023yolo}.
YOLO-v8 is a state-of-the-art object detection model known for its superior real-time performance and accuracy by prioritizing speed; 
while DINO enhances DETR with improved denoising anchor boxes for object detection precision by simplifying the detection process by reducing anchor box reliance.

\noindent \textbf{DINO.} 
In our experiment, the DINO\cite{zhang2022dino} model was fine-tuned for two distinct detectors: one for leaves (1 class) and another for plants (2 classes covering crop and weed).
We experimented with several model architectures and scales from detrex project\cite{ren2023detrex} including DINO-Focal-Large-4scale with 4 focal levels, DINO-ViTDet-Base-4scale and DETA Swin Large.

Our result shows that the model DINO-Focal-Large-4scale (backbone FocalNet-384-LRF-3Level) pre-trained on IN22k gives the best result.

\noindent \textbf{YOLO-v8.} 
Initially, we fine-tuned the YOLO-v8 nano model with four classes (crop, weed, partial crop, and partial weed).
However, this configuration did not result in good performance.
We then decided to fine-tune it with two classes, following the same setup as our DINO fine-tuning.


Interestingly, despite its significantly smaller parameter size than DINO, YOLO-v8 nano performs comparably to DINO in weed detection.
So, we developed a strategy to combine the weed detections from both DINO and YOLO-v8.
This strategy was implemented by removing duplicate bounding boxes using an Intersection over Union (IOU) threshold of 0.5.
Therefore, any additional weed boxes predicted by YOLO-v8 that exceeded 50 pixels in width and height were added to the final detection list.

The implementation of YOLO-v8 all follows default parameters and augmentations from ultralystics\cite{jocher2023yolo}.

\noindent \textbf{Discussion on Object Detection}
During the experiment, we empirically found out that YOLO-v8 performed better in the case of weeds compared to DINO, while DINO showed better performance in detecting leaves and crops.
As such, we utilize both methods as our detectors and employ them to extract the bounding boxes as prompts for SAM as in our pipeline. 

Additionally, in fine-tuning the plant detector, we utilized  Automatic Mixed Precision (AMP) to mitigate memory constraints, given the task involved two classes in contrast to the 1 class for the leaf detector.
During the training of the leaf detector, we observed that using AMP resulted in lower Average Precision (AP) scores compared to training with no-AMP.
This observation highlights the task-specific nature of the impact of AMP, emphasizing the need for careful consideration when deciding to apply this technique in various scenarios.

\begin{table}[h]
\caption{Comparing our result to the other method's results reported in \cite{weyler2023phenobench}.
The best and second best results are respectively highlighted in \textbf{bold}, while "-" means not available.}
\centering
\begin{adjustbox}{width=\linewidth}
\def\arraystretch{1.3}
\begin{tabular}{ccccccc}
\hline
\multirow{2}{*}{\textbf{Methods}} & \multicolumn{6}{c}{\textbf{PhenoBench} \cite{weyler2023phenobench}} \\ \cline{2-7} 
                  & \textbf{IoU (soil)}   & \textbf{IoU (weed)}   & \textbf{PQ (leaf)}   & \textbf{PQ (crop)}   & \textbf{PQ}  & \textbf{PQ+}  \\ \hline
Panoptic DeepLab \cite{cheng2020panoptic}& 99.27  &  - &  -     &  52.02   & -  & 57.97  \\
Mask R-CNN \cite{he2017mask}             & 98.47  &  - & 59.74  &  67.61   & -  & 65.79  \\
Mask2Former \cite{cheng2022masked}       & 98.38  &  - & 57.50  &  71.21   & -  &  69.99 \\
HAPT \cite{roggiolani2023hierarchical}   & 98.50  & 61.11  & 46.84  &  54.61   &  50.73  & 65.27 \\ \hline
\textbf{Ours}                &   \textbf{99.18}     & \textbf{70.66}      & \textbf{73.81}     & \textbf{81.66}     & \textbf{77.73} & \textbf{81.33}  \\ \hline
\end{tabular}
\end{adjustbox}
\label{tab:sota}
\end{table}

\subsection{Experiment Results}
\label{sec:result}

The results of our experiments with various configurations, as reported by the competition's evaluation, are shown in \Cref{tab:experiment_result}.
As in \Cref{tab:experiment_result}, for submission (1) (2), we utilize YOLO-v8 to extract bounding boxes for prompting with 1 class for leaves and four classes for plants and SAM mask decoder for plants. For (3) and (4), the SAM mask decoder predicts plants and leaves, while for the first four submissions, we employed YOLO-v8 nano as the base bounding boxes extractor.
For (5) and (6), we explored the different weights for leaves and plants in the SAM encoder to strengthen the initial feature map by utilizing LORA based mechanism to encoder and decoder on images and masks and also applied DINO FocalNet on four different scales with three different focal levels.

\begin{figure}
     \centering
     \begin{subfigure}[b]{1.0\linewidth}
         \centering
         \includegraphics[width=\linewidth]{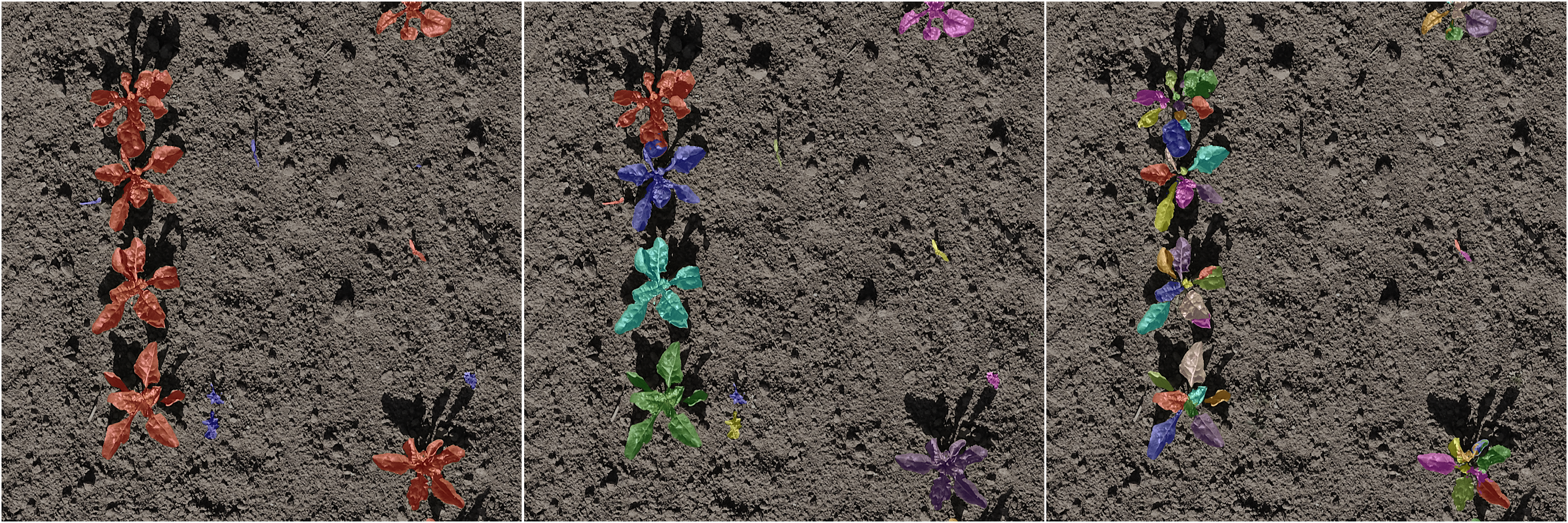}
         \caption{Baseline HQ-SAM with ground-truth bounding box.}
     \end{subfigure}
     \hfill
     \begin{subfigure}[b]{1.0\linewidth}
         \centering
         \includegraphics[width=\linewidth]{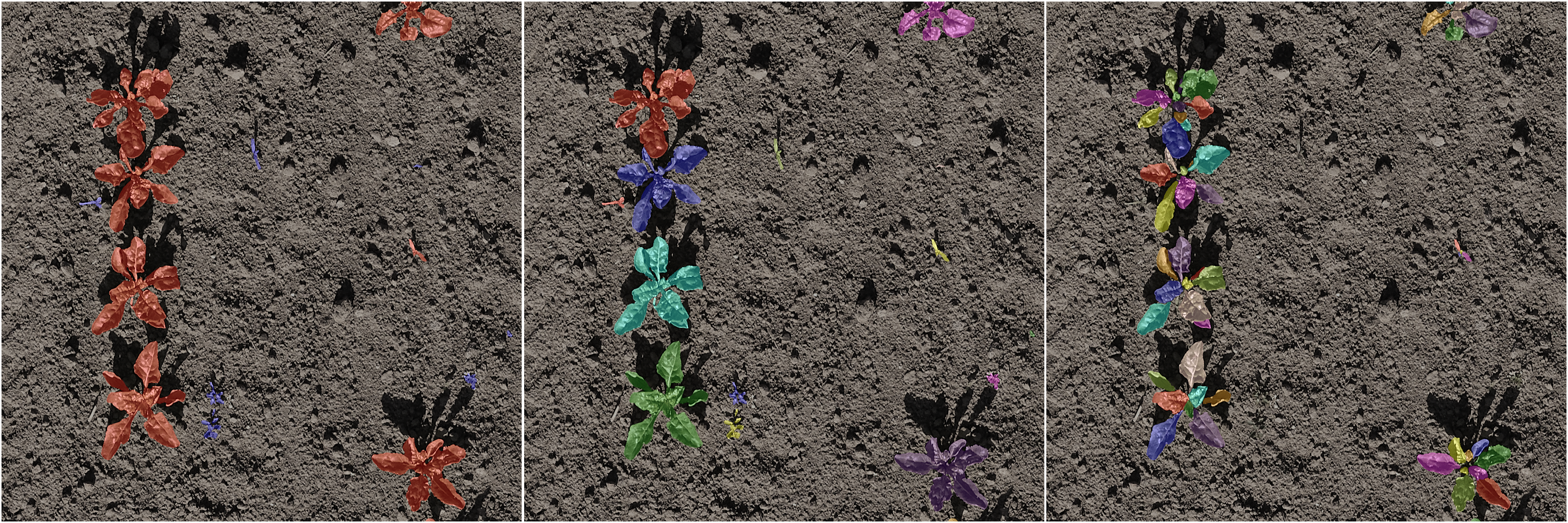}
         \caption{Fine-tuned HQ-SAM with ground-truth bounding box.}
     \end{subfigure}
     \hfill
     \begin{subfigure}[b]{1.0\linewidth}
         \centering
         \includegraphics[width=\linewidth]{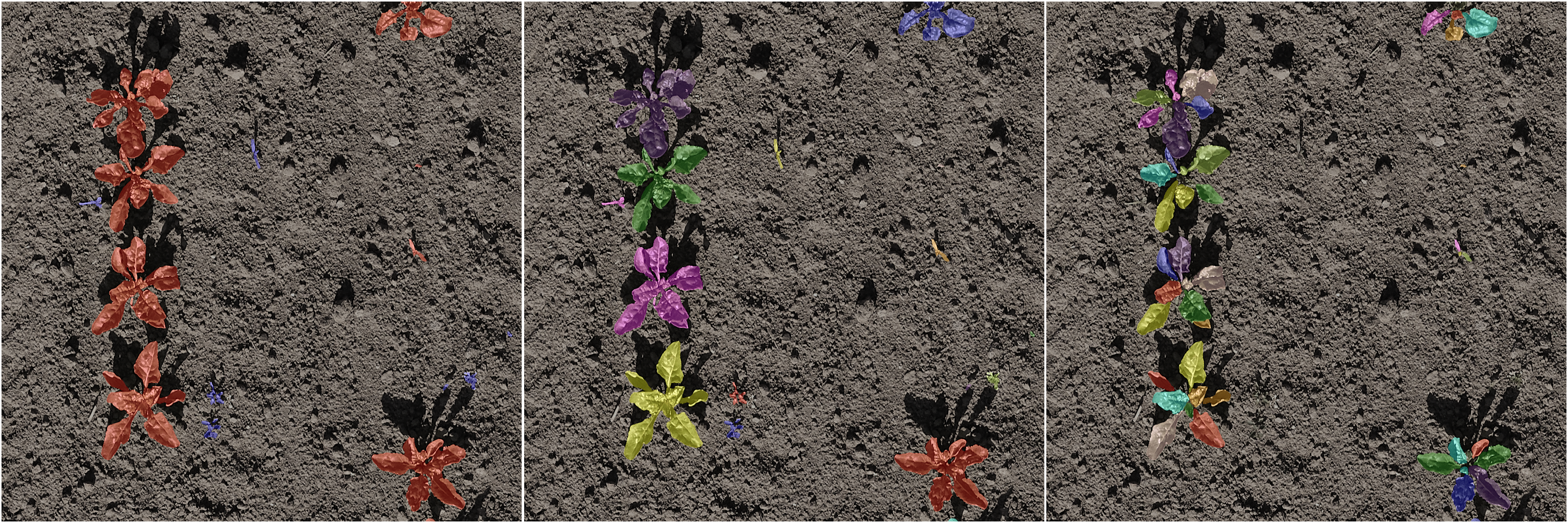}
         \caption{Ours: fine-tuned HQ-SAM with predicted bounding box from YOLO-v8 and DINO.}
         \label{}
     \end{subfigure}
        \caption{Comparison results of different configurations in different tasks, from left to right: semantic segmentation, plant instance segmentation, and leaf instance segmentation}
        \label{fig:example_result}
\end{figure}

\Cref{fig:example_result} shows an example result comparing different configurations of a SAM-based model with ground-truth bounding boxes and bounding boxes generated from the object detection model YOLO-V8 and DINO.
We can observe the effectiveness of the SAM-based model in zero-shot generalization to unfamiliar objects and images.
Even without any fine-tuning, the results of the baseline HQ-SAM with a ground-truth prompt are already reasonable.
Furthermore, after the fine-tuning process, the performance of HQ-SAM is enhanced.
Our approach using fine-tuned HQ-SAM with generated bounding boxes from fine-tuned YOLO-v8 and DINO also performed better than the baseline HQ-SAM.

We also compare our results with other methods reported in \cite{weyler2023phenobench} in \autoref{tab:sota} for reference.

\section{Conclusion}
This report presents our approach to compete at the challenge of Hierarchical Panoptic Segmentation of Crops and Weeds.
Our system is modularized into two separate components: SAM-based segmentation and object detection.
We leverage the capabilities of SAM-based segmentation with generated prompt bounding boxes from the object detection model.
For SAM-based segmentation, we use HQ-SAM, which has an advantage in handling objects with thin lines compared to the original SAM model.
For object detection, we choose DINO and YOLO-v8.
Based on our experimental findings, we observed that YOLO-v8 exhibited notably superior performance in detecting weeds, whereas DINO demonstrated superior proficiency in identifying leaves and crops.
Our best-performing model on PhenoBench achieved a PQ+ score of 81.33 based on the evaluation metrics of the competition.

{\small
\bibliographystyle{ieee_fullname}
\bibliography{egbib}
}

\end{document}